%% file: IWAI2022_TransitionPolicy.tex
\documentclass[runningheads]{llncs}
\usepackage[T1]{fontenc}
\usepackage{graphicx}
\usepackage{url}
\usepackage[utf8]{inputenc}
\usepackage[all]{xy}
\usepackage{todonotes}
\usepackage{array}
\usepackage{listings}
\usepackage{csquotes}
\usepackage{amsmath,amssymb}
\usepackage{float}
\usepackage{hyperref}
\usepackage[nameinlink]{cleveref}
\usepackage{comment}
\usepackage{siunitx}
\usepackage{mathtools}
\usepackage{multicol}
\usepackage{algorithm}
\usepackage{wrapfig}
\usepackage{lipsum}
\usepackage{bm}

\usepackage{fancyhdr}
\usepackage{algpseudocode}

\algdef{SE}[SUBALG]{Indent}{EndIndent}{}{\algorithmicend\ }%
\algtext*{EndIf}
\algtext*{EndFor}
\algtext*{EndLoop}
\algtext*{EndWhile}
\algtext*{Indent}
\algtext*{EndIndent}

\DeclarePairedDelimiterX{\infdivx}[2]{(}{)}{%
  #1\;\delimsize\|\;#2%
}
\newcommand{\kldiv}{KL\infdivx}

\begin{document}

\title{Learning Policies for Continuous Control via Transition Models}

\author{Justus Huebotter\inst{1}\orcidID{0000-0001-8354-0368} \and
Serge Thill\inst{1}\orcidID{0000-0003-1177-4119} \and\\
Marcel van Gerven\inst{1}\orcidID{0000-0002-2206-9098} \and Pablo Lanillos\inst{1}\orcidID{0000-0001-9154-0798}}
\authorrunning{J. Huebotter et al.}
%
\institute{Donders Institute, Radboud University, Nijmegen, The Netherlands
\email{justus.huebotter@donders.ru.nl}}

\maketitle              
\begin{abstract}
It is doubtful that animals have perfect inverse models of their limbs (e.g., what muscle contraction must be applied to every joint to reach a particular location in space). However, in robot control, moving an arm's end-effector to a target position or along a target trajectory requires accurate forward and inverse models. Here we show that by learning the transition (forward) model from interaction, we can use it to drive the learning of an amortized policy. Hence, we revisit policy optimization in relation to the deep active inference framework and describe a modular neural network architecture that simultaneously learns the system dynamics from prediction errors and the stochastic policy that generates suitable continuous control commands to reach a desired reference position. We evaluated the model by comparing it against the baseline of a linear quadratic regulator, and conclude with additional steps to take toward human-like motor control.

\end{abstract}

\section{Introduction}\label{sec:introduction}

Using models for adaptive motor control in artificial agents inspired by neuroscience is a promising road to develop robots that might match human capabilities and flexibility and provides a way to explicitly implement and test these models and its underlying assumptions. 

The use of prediction models in motor planning and control in biological agents has been extensively studied \cite{Krakauer2019,McNamee2019}. Active Inference (AIF) is a mathematical framework that provides a specific explanation to the nature of these predictive models and is getting increased attention from both the neuroscience and machine learning research community, specifically in the domain of embodied artificial intelligence \cite{Lanillos2021,DaCosta2022}. At the core of AIF lies the presence of a powerful generative model that drives perception, control, learning, and planning all based on the same principle of free energy minimization \cite{Friston2011}. However, learning these generative models remains challenging. Recent computational implementations harness the power of neural networks (deep active inference) to solve a variety of tasks based on these principles~\cite{Lanillos2021}. 


While the majority of the state of the art in deep AIF (dAIF) is focused on abstract decision making with discrete actions, in the context of robot control continuous action and state representations are essential, at least at the lowest level of a movement generating hierarchy. Continuous control implementations of AIF, based on the original work from Friston~\cite{Friston2011}, is very well suited for adaptation to external perturbations~\cite{Sancaktar2020} but it computes suboptimal trajectories and enforces the state estimation to be biased to the preference/target state~\cite{Lanillos2021}. New planning algorithms based on optimizing the expected free energy~\cite{Millidge2021} finally uncouple the action plan from the estimation but they suffer from complications to learn the generative model and the preferences, specially for generating the actions.

In this paper, we revisit policy optimization using neural networks from the perspective of predictive control to learn a low-level controller for a reaching task. We show that by learning the transition (forward) model, during interaction, we can use it to drive the learning of an amortized policy. The proposed methods are not entirely novel, but instead combine aspects of various previous methods for low-level continuous control, active inference, and (deep) reinforcement learning. This is an early state proof-of-concept study aimed at understanding how prediction networks can lead to successful action policies, specifically for motor control and robotic tasks.

First, we summarize important related research and then go on to describe a modular neural network architecture that simultaneously learns the system dynamics from prediction errors and the stochastic policy that generates suitable continuous control commands to reach a desired reference position. Finally, we evaluated the model by comparing it against the baseline of a linear quadratic regulator (LQR) in a reaching task, and conclude with additional steps to take towards human-like motor control.


\section{Related Work}\label{sec:relatedwork}

This work revisits continuous control and motor learning in combination with system identification, an active direction of research with many theoretical influences. As the body of literature covering this domain is extensive, a complete list of theoretical implications and implementation attempts goes beyond the scope of this paper. Instead, we want to highlight selected examples that either represent a branch of research well or have particularly relevant ideas.

Motor learning and adaptation has been studied extensively in humans (for recent reviews please see \cite{Krakauer2019,McNamee2019}). Humans show highly adaptive behavior to perturbations in simple reaching tasks and we aim to reproduce these capabilities in artificial agents. While simple motor control can be implemented via optimal control when the task dynamics are known \cite{Krakauer2019}, systems that both have learning from experience and adaptation to changes have had little attention~\cite{DeWolf2016,Bian2020}. However, the assumption that the full forward and inverse model are given is not often met in practice and hence these have to be learned from experience~\cite{wolpert1998multiple}. Initial experiments in reaching tasks for online learning of robot arm dynamics in spiking neural networks inspired by optimal control theory have shown promising results~\cite{Iacob2020}.

Recently, the most dominant method for control of unspecified systems in machine learning is likely that of deep reinforcement learning (dRL) where control is learned as amortized inference in neural networks which seek to maximize cumulative reward. The model of the agent and task dynamics is learned either implicitly (model-free) \cite{Lee2020} or explicitly (model-based) \cite{Hafner2019dream,Wu2022daydreamer} from experience. The advantage of an explicit generative world model is that it can be used for planning \cite{Traub2021}, related to model predictive control, or generating training data via imagining \cite{Hafner2019dream,Wu2022daydreamer}. Learning and updating such world models, however, can be comparatively expensive and slow. Recently, there has been a development towards hybrid methods that combine the asymptotic performance of model-free with the planning capabilities of model-based approaches \cite{Tschantz2020}. Finally, model-free online learning for fast motor adaptation when an internal model is inaccurate or unavailable \cite{Bian2020} shows promising results that are in line with behavioral findings in human experiments and can account for previously inexplicable key phenomena.


The idea of utilizing a generative model of the world is a core component of AIF, a framework unifying perception, planning, and action by jointly minimizing the expected free energy (EFE) of the agent \cite{Adams2013,Friston2011,Lanillos2021}. In fact, here this generative model entirely replaces the need for an inverse model (or policy model in RL terms), as the forward model within the hierarchical generative model can be inverted directly by the means of predictive coding. This understands action as a process of iterative, not amortized, inference and is hence a strong contrast to optimal control theory, which requires both forward and inverse models \cite{kalman1960contributions}. Additionally, the notion of exploration across unseen states and actions is included naturally as the free energy notation includes surprise (entropy) minimization, a notion which is artificially added to many modern RL implementations \cite{Hafner2019dream,Wu2022daydreamer,Lee2020}. Also, AIF includes the notion of a global prior over preferred states which is arguably more flexible than the reward seeking of RL agents, as it can be obtained via rewards as well as other methods such as expert imitation. Recently, the idea of unidirectional flow of top-down predictions and bottom-up prediction errors has been challenged by new hybrid predictive coding, which extends these ideas by further adding bottom-up (amortized) inference to the mix \cite{Tschantz2022}, postulating a potential paradigm shift towards learned habitual inverse models of action.

Recent proof-of-concept AIF implementations have shown that this framework is capable of adaptive control, e.g. in robotic arms \cite{Oliver2022} via predictive processing. In practice, most implementations of AIF by the machine learning community use neural networks to learn approximations of the probabilistic quantities relevant in the minimization of the EFE, named deep active inference. Using gradient decent based learning, these forward models can be used to directly propagate the gradients of desired states with respect to the control signals (or policy) \cite{Hafner2019dream,Wu2022daydreamer,Catal2019,Catal2020,vanderHimst2020,Millidge2020}. Input to such policies is commonly given as either fully observable internal variables (related to proprioception) \cite{Ueltzhoeffer2018,Catal2019,Catal2020}, visual observations directly \cite{Lee2020} or a learned latent representation of single \cite{vanderHimst2020,Hafner2019dream,Wu2022daydreamer} or mixed sensory input \cite{Meo2021,Sancaktar2020}. This, however, makes use of amortized inference with bottom-up perception and top-down control \cite{Ueltzhoeffer2018,Catal2019,Catal2020,vanderHimst2020,Millidge2020} and is hence in some contrast to the predictive nature of the original AIF theory and more closely related to deep RL.

In summary, AIF postulates a promising approach to biologically plausible motor control \cite{Friston2011,Adams2013}, specifically for robotic applications \cite{DaCosta2022}. The minimization of an agent's free energy is closely related to other neuroscientific theories such as the Bayesian brain hypothesis and predictive coding. Adaptive models can be readily implemented when system dynamics are known \cite{piolopez2016,DeWolf2016}. Unknown generative models of (forward and, if needed, inverse) dynamics may be learned from various perceptive stimuli through experience in neural networks via back propagation or error \cite{Hafner2019dream,Wu2022daydreamer,Catal2019,Catal2020,vanderHimst2020,Millidge2020,Tschantz2020} or alternative learning methods \cite{Ueltzhoeffer2018,Tschantz2022,Iacob2020}. This can be extended to also learn priors about preferred states and actions \cite{Hafner2019dream,Wu2022daydreamer,Lee2020,Tschantz2020,vanderHimst2020,Catal2019,Catal2020}. Generative models (and their priors) can then be utilized for perception, action, planning \cite{vanderHimst2020,Traub2021}, and the generation of imagined training data \cite{Hafner2019dream,Wu2022daydreamer}.

In this work, we draw inspiration from these recent works. We are learning a generative model for a low-level controller with unknown dynamics from fully observable states through interaction. One component learns the state transitions, which in turn, similar to \cite{Hafner2019dream,Wu2022daydreamer}, is used to generate imagined training data for an amortized policy network. 
The prior about preferred states is assumed to be given to this low-level model and hence no reward based learning is applied.

\section{Model}\label{sec:model}

\begin{figure}
    \centering
    \includegraphics[width=.49\textwidth]{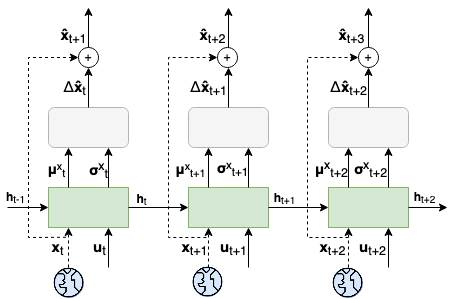}
    \includegraphics[width=.49\textwidth]{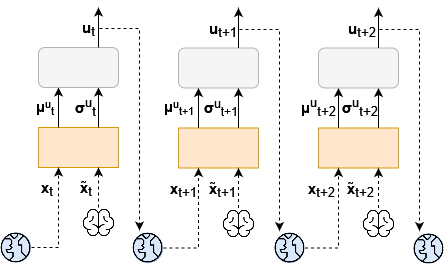}
    \caption{Transition model (left) and policy model (right) workflow over three time steps. The policy network (orange) takes a state $\bm{x}$ and target $\tilde{\bm{x}}$ as input from external sources to generate a control action $\bm{u}$. The recurrent transition network (green) predicts the change to the next state $\Delta\bm{x}$ based on state $\bm{x}$ and control $\bm{u}$. The gray box is a Gaussian sampling process.}
    \label{fig:models}
\end{figure}

We consider a fully observable but noisy system with unknown dynamics. We formalize this system as an Markov Decision Process (MDP) in discrete time $t \in \mathbb{Z}$. The state of the system as an $n$-dimensional vector of continuous variables $\bm{x}_t \in \mathbb{R}^n$. Likewise, we can exert $m$-dimensional control on the system via continuous actions $\bm{u}_t \in \mathbb{R}^m$. We aim to learn a policy that can bring the system to a desired goal state $\tilde{\bm{x}} \in \mathbb{R}^n$, which is assumed to be provided by an external source. If the system dynamics were known, we could apply optimal control theory to find $\bm{u}^*_t$ for each point in time $t \in [0, \infty)$. However, the system dynamics are unknown and have to be learned (system identification). The dynamics of the system are learned via interaction and from prediction errors by a transition model $\upsilon$. This transition model is used to train in parallel a policy model $\pi$ to generate the control actions. Both models are schematically summarized in \Cref{fig:models}. 

\subsection{Transition Model}
The dynamics of the system are described by
\begin{equation}
    \bm{x}_{t+1} = \bm{x}_{t} + f(\bm{x}_t, \ \bm{u}_t, \ \bm{\zeta}_t),
\end{equation}
where $\zeta$ is some unknown process noise. Further, any observation $y$ cannot be assumed to be noiseless and thus
\begin{equation}
\label{eq:observation_noise}
    \bm{y}_t = \bm{x}_t + \bm{\xi}_t,
\end{equation}
where $\xi$ is some unknown observation noise. As $f$ is unknown, we want to learn a function $g$ that can approximate it as
\begin{equation}
    g(\bm{y}_t, \ \bm{u}_t, \ \phi) \approx f(\bm{x}_t, \ \bm{u}_t, \ \bm{\zeta}_t),
\end{equation}
by optimizing the function parameters $\phi$. We hence define a state estimate $\hat{\bm{x}}$ as
\begin{equation}
\label{eq:state_sampling}
    \hat{\bm{x}}_t \sim \mathcal{N}(\bm{\hat{\mu}}^x_t, \ \bm{\hat{\sigma}}^x_t),
\end{equation}
where the superscript $x$ indicates not an exponent but association to the state estimate and
\begin{equation}
    \bm{\hat{\mu}}^x_t = \bm{y}_{t-1} + \bm{\hat{\mu}}^{\Delta x}_t.
\end{equation}
In turn, both $\bm{\hat{\mu}}^{\Delta x}_t$ and $\hat{\bm{\sigma}}^{x}_t = \hat{\bm{\sigma}}^{\Delta x}_t$ are outputs of a learned recurrent neural network (transition network) with parameters $\phi$ as 
\begin{equation}
\label{eq:transition_network}
    \bm{\hat{\mu}}^{\Delta x}_{t}, \ \bm{\hat{\sigma}}^{\Delta x}_t = g(\bm{y}_{t-1}, \ \bm{u}_{t-1}, \ \phi).
\end{equation}
To maintain differentiability to the state estimate we apply the reparametrization trick in \Cref{eq:state_sampling}. Further, we summarize the steps from \Cref{eq:state_sampling} - \ref{eq:transition_network} (the transition model $\upsilon$, see \Cref{fig:models} left) as
\begin{equation}
    \bm{\hat{x}}_t = \upsilon(\bm{y}_{t-1}, \ \bm{u}_{t-1}, \ \phi).
\end{equation}
The optimal transition function parameters $\phi^*$ are given by minimizing the Gaussian negative log-likelihood loss
\begin{equation}
    \mathcal{L}_\upsilon = \frac{1}{2T} \sum_{t=1}^T \left(\log\left(\text{max}\left(\hat{\bm{\sigma}}^x_t, \ \epsilon\right)\right) + \frac{\left(\hat{\bm{\mu}}^{x}_t - \bm{y}_t\right)^2}
        {\text{max}\left(\hat{\bm{\sigma}}^x_t, \ \epsilon\right)}\right),
\end{equation}
and
\begin{equation}
    \phi^* = \operatorname*{argmin}_\phi \mathcal{L}_\upsilon,
\end{equation}
where $\epsilon$ is a small constant to avoid division by zero and the added constant has been omitted.

\subsection{Policy Model} 
The actor is given by the policy $\pi_{\theta}$ that gives a control action $\bm{u}$ for a given current state $\bm{x}$ and target or preferred state $\tilde{\bm{x}}$ 
as 
\begin{equation}
    \pi(\bm{u}_t \ | \ \bm{x}_t, \ \tilde{\bm{x}}_t, \theta),
\end{equation}
where $\bm{x}_t$ can be either an observation from the environment $\bm{y}_t$ or an estimate from the transition network $\hat{\bm{x}}_t$ and
\begin{equation}
    \bm{u}_t \sim \mathcal{N}(\bm{\mu}^u_t, \ \bm{\sigma}^u_t).
\end{equation}
Here, $\bm{\mu}^u$ and $\bm{\sigma}^u$ are given by a function approximator that is a neural network with parameters $\theta$ (see \Cref{fig:models} right). We aim to find the optimal policy $\pi^*$ so that 
\begin{equation}
    \pi^* = \operatorname*{argmin}_u \sum_{t=1}^T \left( \bm{x}_t - \tilde{\bm{x}}_t \right)^2.
\end{equation}
However, as $\bm{x}_t$ is non-differentiable with respect to the action, we instead use the transition model estimate $\hat{\bm{x}}_t$. 
This also allows to find the gradient of the above loss with respect to the action $u$ by using backpropagation through the transition network and the reparametrization trick. Policy and transition network are optimized by two separate optimizers as to avoid that the policy loss pushes the transition network to predict states that are the target state, which would yield wrong results.

While the above formulation in principle should find a system that is able to minimize the distance between the current state estimate $\hat{\bm{x}}$ and the target $\tilde{\bm{x}}$, in practice there are some additional steps to be taken into account to learn a suitable policy. As the state contains information about position and velocity, so does the target state. If the target state is a fixed position, the target velocity is given as zero. However, optimizing the system in a matter where the loss increases as the system starts moving, there is a strong gradient towards performing no action at all, even if this means that the position error will remain large throughout the temporal trajectory. To overcome this issue, we introduce a target gain vector $\tilde{\bm{x}}_g$, which weighs the relevance of each preference state variable. For instance, when the velocity of the system is non-important we set to 1 where $x$ is a representing a position encoding and 0 for every velocity. The weighted policy loss becomes:
\begin{equation}
    \mathcal{L_\pi} = \frac{1}{T} \sum_{t=1}^T \tilde{\bm{x}}_g \left( \hat{\bm{x}}_t - \tilde{\bm{x}} \right)^2.
\end{equation}

The offline training procedure for both transition and policy networks is summarized in algorithm 1 below, as well as \Cref{alg:transition_offline} \& \ref{alg:policy_offline} in the \Cref{sec:algorithms} and \ref{sec:parameters}.

\begin{algorithm}[H]
\label{alg:offline_training}
\caption{Offline training of transition and policy networks}
\begin{algorithmic}[1]
	\State
	Input: a differentiable transition parametrization $\upsilon(\hat{\bm{x}}'|\bm{y},\bm{u},\bm{\phi})$,\\
	a differentiable policy parametrization $\pi (\bm{u}|\bm{x},\tilde{\bm{x}},\bm{\theta})$,\\
	a task environment providing $(\bm{y}',\tilde{\bm{x}}'|\bm{u})$\\
	Initialize transition parameters $\bm{\phi} \in \mathbb{R}^d$ and policy
	parameters $\bm{\theta} \in \mathbb{R}^{d'}$\\
	Initialize a memory buffer of capacity $M$
	\Loop\, for $I$ iterations:
	    \State Play out $E$ episodes of length $T$ by applying $\bm{u} \sim \pi(\bm{y}, \tilde{\bm{x}},\bm{\theta})$ at each step and save to memory
	    \State Update transition network parameters for $n_{\upsilon}$ batches of size $N_{\upsilon}$ sampled from memory
	    \State Update policy network parameters for $n_{\pi}$ batches of size $N_{\pi}$ sampled from memory
    \EndLoop
\end{algorithmic}
\end{algorithm}

\section{Results}\label{sec:results}

\begin{wrapfigure}{R}{5cm}
    \vspace{-2cm}
    \centering
    \includegraphics[width=0.2\textwidth]{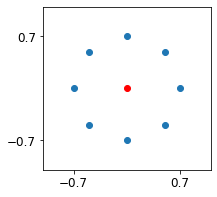}
    \caption{Eight equidistant targets (blue) are presented to the agent in sequence, starting from the center position (red) each time.}
    \label{fig:eval_tasks}
\end{wrapfigure}

Here we summarize the key results of this research. For a more detailed description of the task please refer to appendix \Cref{sec:task}. To evaluate the performance of the trained models in comparison to an LQR baseline we have established a reaching task inspired by experiments conducted in humans and robots in previous research \cite{Krakauer2019,DeWolf2016}. The agent is presented eight equidistant targets in sequence for $T = 200$ steps, while starting at the center position $\bm{x}_{t0} = [0, 0, 0, 0]$. Initially, each target is 0.7 units of distance removed from the center with offsets of $45^{\circ}$ (\Cref{fig:eval_tasks}). In one case these targets are stationary, or alternatively rotate in a clockwise motion with an initial velocity of 0.5 perpendicular to the center-pointing vector.
To test agent performance under changed task dynamics, we offset the rotation angle $\gamma$ during some evaluations, which influences the direction of the acceleration as given by control $\bm{u}$ (see \Cref{eq:rotation}). 
To quantify the performance of target reaching, we measure the Euclidean distance between the current position $[x_1, x_2]$ and the target position $[\tilde{x}_1, \tilde{x}_2]$ at each step $t$, so that performance is defined as

\begin{equation}
    J = \Delta t \sum_{t=1}^T \sqrt{(x_{1,t} - \tilde{x}_{1,t})^2 + (x_{2,t} - \tilde{x}_{2,t})^2}\,.
\end{equation}

\begin{figure}
    \centering
    \includegraphics[width=0.4\textwidth]{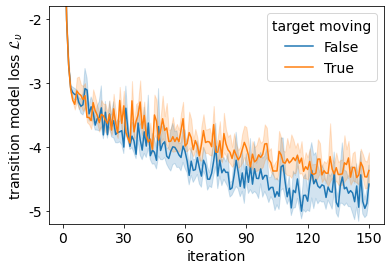}
    \includegraphics[width=0.4\textwidth]{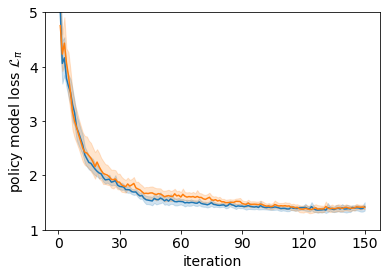}
    \includegraphics[width=0.4\textwidth]{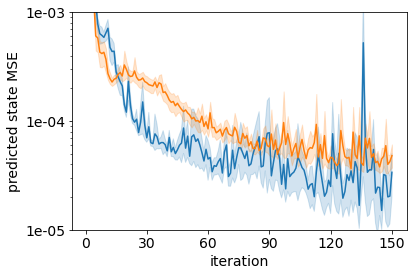}
    \includegraphics[width=0.4\textwidth]{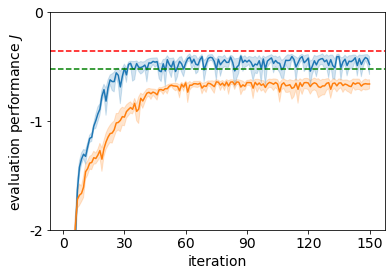}
    \caption{Model performance improves during learning. The transition model shows better predictions when the target is stationary. The policy closely approaches but never reaches the LQR baseline scores for both stationary (red dotted line) and moving targets (green dotted line).}
    \label{fig:results}
\end{figure}

Results in \Cref{fig:results} show that both the transition model and policy model are able to quickly learn from the environment interactions. The task of reaching stationary targets only is easier to conduct with predicted state mean squared error lower and a higher evaluation task performance. For both tasks, the model performance approached but never fully reached the optimal control baseline of LQR -- for implementation details of the baseline please refer to appendix \Cref{sec:lqr}).

\begin{figure}[!htb]
    \centering
    \includegraphics[width=0.32\textwidth]{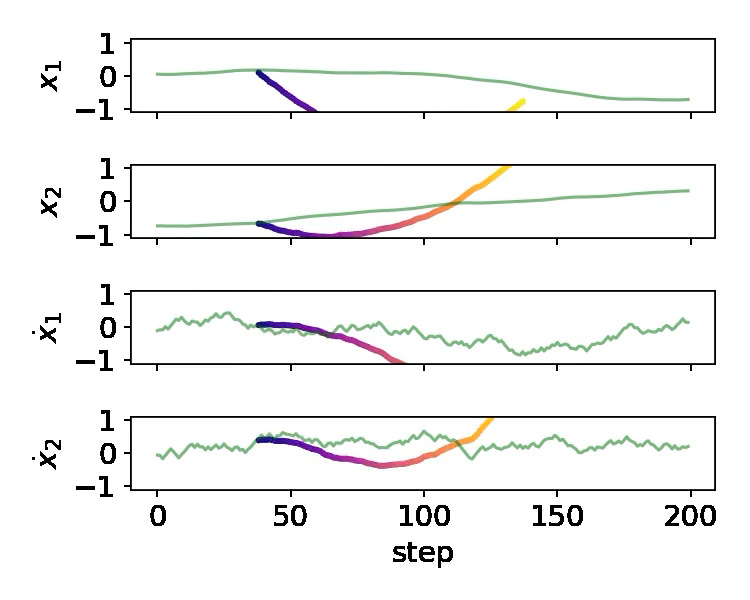}
    \includegraphics[width=0.32\textwidth]{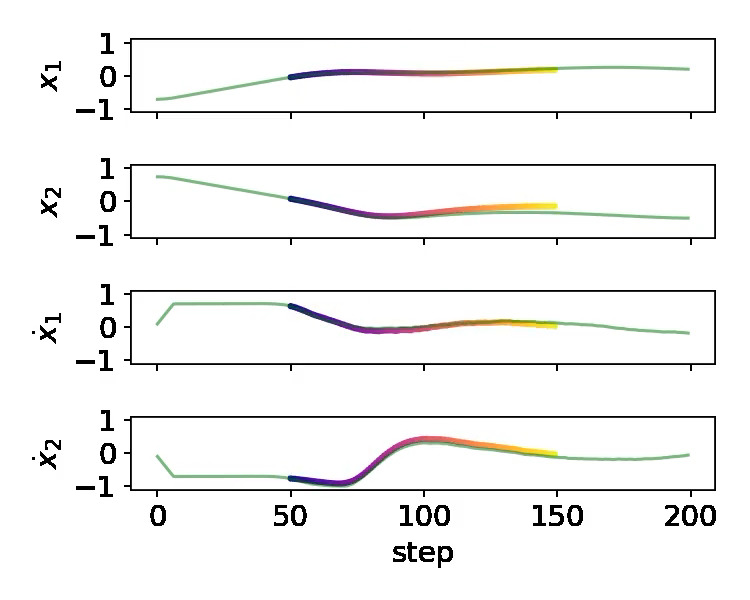}
    \includegraphics[width=0.32\textwidth]{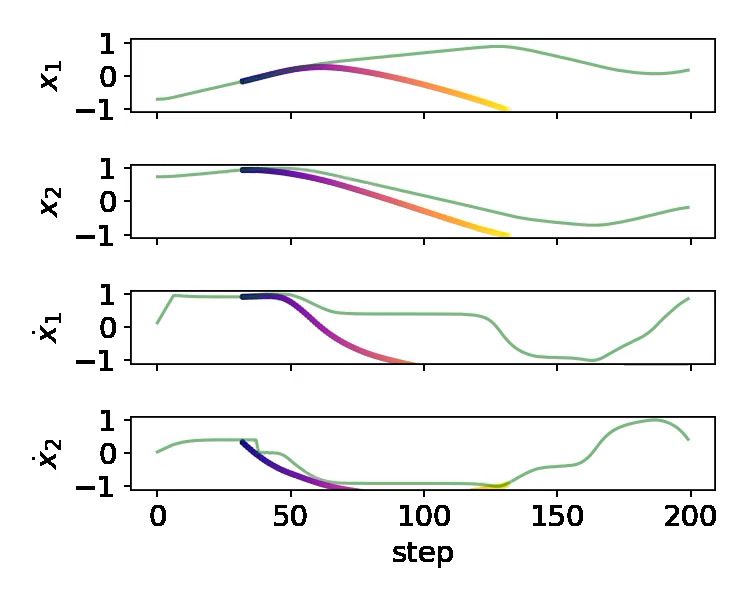}
    \caption{Auto-regressive transition model predictions (blue to yellow) for 100 time steps over the true state development (green) are poor at the beginning of training (left), but can closely follow the true state development at the end of the training (center). Perturbing the action with a rotation angle $\gamma = 60^{\circ}$ induces a mismatch between state prediction and true trajectory (right).} 
    \label{fig:predictions}
\end{figure}

\Cref{fig:predictions} shows auto-regressive predictions of the transition model when provided with some initial states and the future action trajectory. The model initially failed to make sensible predictions, but the final trained model closely predicts the true state development. When applying a rotational perturbation to the input control of $\gamma=60^{\circ}$ (\Cref{fig:predictions}(right)) these predictions start to diverge from the true state, as the model has capabilities for online adaptation. 

\begin{figure}
    \centering
    \begin{tabular}{c c c c c}
        \begin{tabular}{l}
             A\\
             \includegraphics[width=0.16\textwidth]{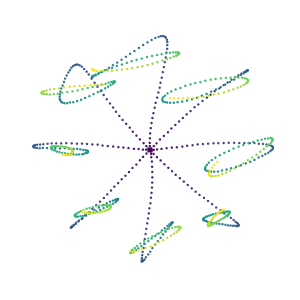} \\
             E\\
             \includegraphics[width=0.16\textwidth]{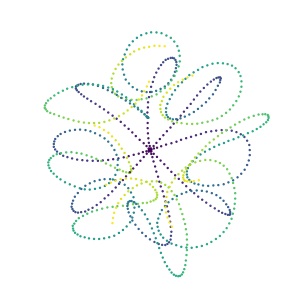}
        \end{tabular}
        &
        \begin{tabular}{l}
             B\\
             \includegraphics[width=0.16\textwidth]{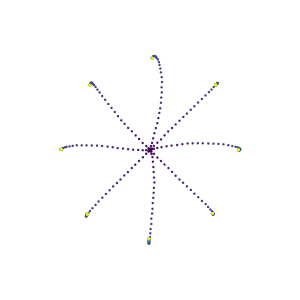}\\
             F\\
             \includegraphics[width=0.16\textwidth]{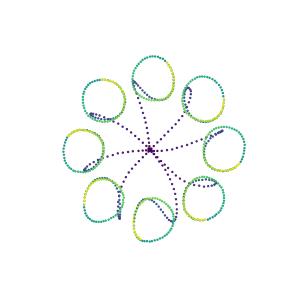}
        \end{tabular}
        & 
        \begin{tabular}{l}
             C\\
             \includegraphics[width=0.16\textwidth]{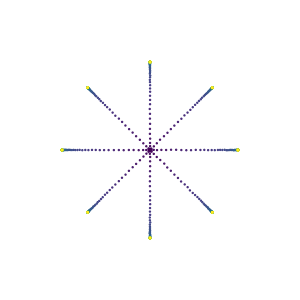} \\
             G\\
             \includegraphics[width=0.16\textwidth]{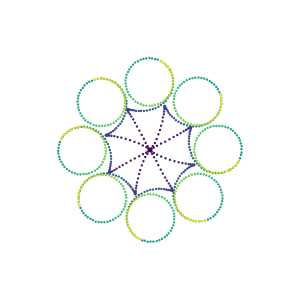}
        \end{tabular}
        &
        \begin{tabular}{l}
            D\\
            \includegraphics[width=0.16\textwidth]{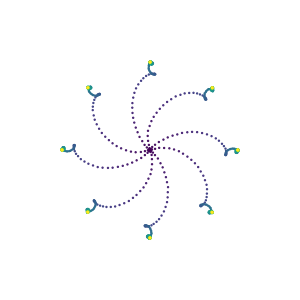}\\
            H\\
            \includegraphics[width=0.16\textwidth]{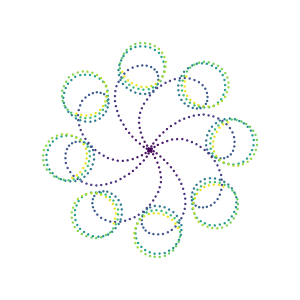}
        \end{tabular}
        &
        \begin{tabular}{c}
             \includegraphics[width=.3\textwidth]{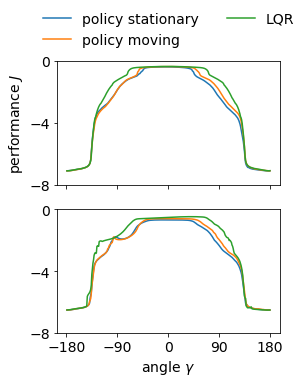}
        \end{tabular}
    \end{tabular}
    \caption{Example trajectory plots from the evaluation task for stationary targets (top row) and moving targets (bottom row) show the improvement during learning from iteration 20 (A \& E) to iteration 150 (B \& F). The LQR baseline performs the target reaching optimally (C \& D), but struggles when the control input $u$ is rotated by $\gamma = 60^{\circ}$ (D \& H). The graph on the right shows that both models learned on stationary as well as moving targets perform close to the LQR under different perturbation conditions, but no model can reach the targets when the rotation becomes larger than $\gamma = 90^{\circ}$.}
    \label{fig:trajectories}
\end{figure}

The policy model is initially unable to complete the reaching task, but has a strong directional bias of movement (data not shown). After just 20 iterations (200 played episodes and 600 policy weight updates) we observe that the policy model can partially solve target reaching for both stationary and moving targets (\Cref{fig:trajectories} A \& E respectively). At the end of training the model generated trajectories (B \& F) closely match those of the LQR baseline (C \& G). Applying perturbations results in non-optimal trajectories to the target (D \& H). Once these perturbations become too large at around $\gamma = \pm 90^{\circ}$, neither LQR nor the learned models can solve the tasks. However, the learned models closely track the performance of the LQR. This failure is a result of both policy and transition model being learned entirely offline and the inference being completely amortized. We believe that a more predictive coding based implementation of AIF as suggested by \cite{Adams2013,Friston2011} and demonstrated by \cite{piolopez2016} would allow the system to recover from such perturbations. In future iterations of this research, we aim to extend both the transition and policy models by an adaptive component that can learn online from prediction errors to recover performance similar to \cite{DeWolf2016,Iacob2020} and match adaptation similar to that described in humans \cite{Krakauer2019}. 

\section{Conclusion}\label{sec:conclusion}

Here, we show that a low-level motor controller and its state dynamics can be learned directly from prediction error via offline learning. Furthermore, it has similar capabilities to LQR to absorb rototranslation perturbations. However, as neither model has any means of online adaptation, they fail to show the behavioral changes described in humans \cite{Krakauer2019} or control approaches \cite{DeWolf2016}. In future research, hope to take steps towards human-like online motor adaptation as described in \cite{Krakauer2019,McNamee2019}. AIF proposes a specific implementation of prediction error-driven motor action generation \cite{Friston2011,Adams2013,piolopez2016}, but computational implementations in dRL and dAIF based on offline learning in neural networks often lack these online adaptation capabilities. In future iterations of this research, we aim to address this gap . Specifically, we propose to combine the offline learning of our model with model-free adaptation, such as e.g. presented in \cite{Bian2020,DeWolf2016}.


Our implementation is based on some underlying assumptions. There are two kinds of input to the system that come from other components of a cognitive agent which we do not explicitly model. First, the position of the agent effector in relation to some reference frame (e.g. its base joint) is provided to the low-level controller in Cartesian coordinates. This information would have to be obtained through an integration of visual, proprioceptive, and touch information. Second, the target position of this effector is provided in the same coordinate system. This information would likely be generated by a motor planning area where abstract, discrete action priors (e.g. grasp an object) are broken down into a temporal sequence of target positions. Integrating our method with models of these particular systems is not part of this work but should be addressed in future research.

\subsubsection*{Acknowledgements}

This research was partially funded by the Human Brain Project SGA3.

\vfill
\pagebreak

\bibliographystyle{splncs04}
\bibliography{IWAI2022_TransitionPolicy}

\vfill
\pagebreak

\appendix
\input{appendix}

\end{document}

%% file: appendix.tex
\newpage
\section*{Appendix}
\label{sec:appendix}

\section{Task Description}
\label{sec:task}

The state of the 2d plane environment is given as
\begin{equation}
    \bm{x} = [x_1, x_2, \dot{x}_1, \dot{x}_2].
\end{equation}
Further, the desired target state is given as
\begin{equation}
    \tilde{\bm{x}} = [\tilde{x}_1, \tilde{x}_2, \tilde{\dot{x}}_1, \tilde{\dot{x}}_2].
\end{equation}
When we only care about the final position in the state, then the target gain is
\begin{equation}
    \tilde{\bm{x}}_g = [\tilde{x}_{g1}, \tilde{x}_{g2}, \tilde{\dot{x}}_{g1}, \tilde{\dot{x}}_{g2}] = [1, 1, 0, 0].
\end{equation}

The desired target state as well as it's target gain are currently provided by the task itself, but later should be provided by some higher level cognitive mechanism.

Further, the action influences the state by
\begin{equation}
    \bm{u} = [u_1,\, u_2] \propto [\ddot{x}_1, \ddot{x}_2],
\end{equation}
where $u_i \in [- u_{max}, u_{max}]$.

Following the forward Euler for discrete time steps with step size $\Delta t$ we also get the environment dynamics as
\begin{equation}
\label{eq:position}
    \hat{x}_{i,t+1} \sim \mathcal{N}(x_{i,t} + \Delta t\dot{x}_{i,t},\,\zeta_x),
\end{equation}
and then clip the computed value based on the constrains
\begin{equation}
    x_{i,t+1} = \begin{cases}
        x_{max} & \text{if}\, \hat{x}_{i,t+1} > x_{max}\\
        \hat{x}_{i,t+1} & \text{if}\, x_{max} > \hat{x}_{i,t+1} > x_{min}\\
        x_{min} & \text{if}\, \hat{x}_{i,t+1} < x_{min}
                \end{cases}
\end{equation}
Doing the same for velocity and acceleration we get
\begin{equation}
\label{eq:velocity}
    \hat{\dot{x}}_{i,t+1} \sim \mathcal{N}(\dot{x}_{i,t} + \Delta t\ddot{x}_{i,t},\,\zeta_{\dot{x}}),
\end{equation}
and
\begin{equation}
    \dot{x}_{i,t+1} = \begin{cases}
        \dot{x}_{max} & \text{if}\, \hat{\dot{x}}_{i,t+1} > \dot{x}_{max}\\
        \hat{\dot{x}}_{i,t+1} & \text{if}\, \dot{x}_{max} > \hat{\dot{x}}_{i,t+1} > \dot{x}_{min}\\
        \dot{x}_{min} & \text{if}\, \hat{\dot{x}}_{i,t+1} < \dot{x}_{min}
                \end{cases}
\end{equation}
as well as
\begin{equation}
\label{eq:acceleration}
    \hat{\ddot{x}}_{i,t+1} \sim \mathcal{N}(\kappa u_{i,t}',\,\zeta_{\ddot{x}}),
\end{equation}
where $\kappa$ is some real valued action gain and $u'$ may be subject to a rotation by the angle $\gamma$ as
\begin{equation}
\label{eq:rotation}
    \bm{u'} = \bm{u} * \begin{bmatrix}
        \cos{\gamma}, -\sin{\gamma}\\
        \sin{\gamma}, \cos{\gamma}
    \end{bmatrix}.
\end{equation}
Finally,
\begin{equation}
    \ddot{x}_{i,t+1} = \begin{cases}
        \ddot{x}_{max} & \text{if}\, \hat{\ddot{x}}_{i,t+1} > \ddot{x}_{max}\\
        \hat{\ddot{x}}_{i,t+1} & \text{if}\, \ddot{x}_{max} > \hat{\ddot{x}}_{i,t+1} > \ddot{x}_{min}\\
        \ddot{x}_{min} & \text{if}\, \hat{\ddot{x}}_{i,t+1} < \ddot{x}_{min}
                \end{cases}
\end{equation}
where $\bm{\zeta} = [\zeta_x,\, \zeta_{\dot{x}},\, \zeta_{\ddot{x}}]$ is some Gaussian process noise parameter and the maximum and minimum values are the boundaries of space, velocity, and acceleration respectively. In the normal case $\zeta=[0,\,0,\,0]$, so that there is no process noise unless explicitly mentioned otherwise. Here, we can see that updating the state $\bm{x}$ by following \Cref{eq:position} to \Cref{eq:acceleration} in this order, it takes three steps for any control signal to have an effect on the position of the agent itself. This is why it is necessary to use a RNN as the transition model to grasp the full relationship between control input and state dynamics.

Finally, the environment adds some observation noise $\bm{\xi} = [\xi_x,\, \xi_{\dot{x}}]$ to the state before providing it back to the controller, as mentioned in \Cref{eq:observation_noise}, so that
\begin{equation}
    \bm{y} = [y_1, y_2, \dot{y}_1, \dot{y}_2],
\end{equation}
with
\begin{equation}
    y_{i,t} \sim \mathcal{N}(x_{i,t},\, \xi_x),
\end{equation}
\begin{equation}
    \dot{y}_{i,t} \sim \mathcal{N}(\dot{x}_{i,t},\, \xi_{\dot{x}}).
\end{equation}

\vfill
\pagebreak

\section{Training Algorithms}
\label{sec:algorithms}

\begin{figure}[H]
    \centering
    \includegraphics[width=.48\textwidth]{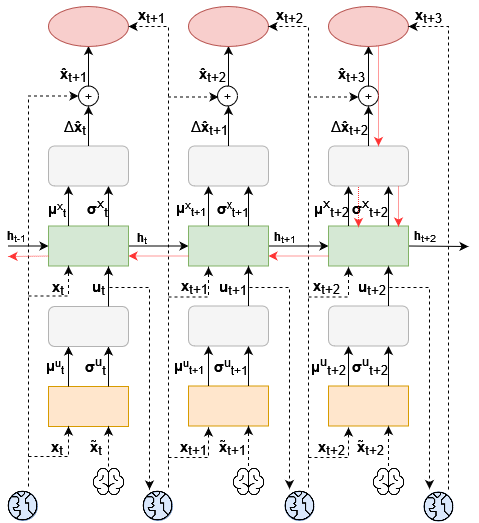}
    \hfill
    \includegraphics[width=.48\textwidth]{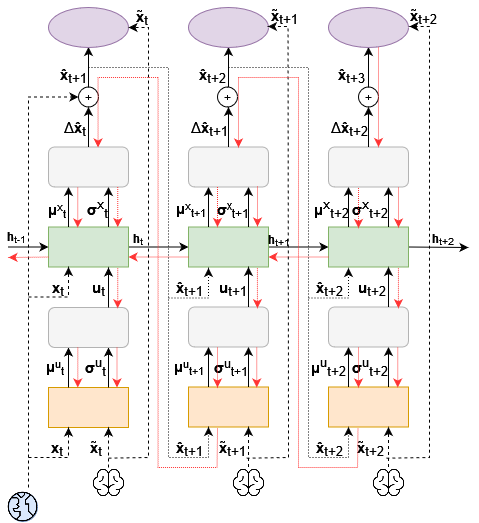}
    \caption{Transition model learning (left) and policy model learning (right) use different algorithms. The transition model directly tries to predict the change in state and the gradients (red arrows) can directly flow from the loss computation (red) through the sampling step (gray) and to the recurrent model parameters (green). In case of the policy model update, the procedure is more involved. In order to obtain gradients with respect to the action, the models jointly roll out an imagined state and action sequence in an auto-regressive manner. The gradients have to flow from its own loss function (purple) through the transition model to reach the policy parameters (orange). This assumes that the transition model is sufficiently good at approximating the system dynamics.}
    \label{fig:model_learning}
\end{figure}

The following two algorithms describe in more detail the offline learning of the transition network (algorithm 2) and policy network (algorithm 3) that correspond to lines 8 and 9 of algorithm 1 respectively. For a summary please refer to \Cref{fig:model_learning}).

\begin{algorithm}[H]
\caption{Updating of transition network parameters}
\begin{algorithmic}[1]
\label{alg:transition_offline}
	\State
	Input: a differentiable transition parametrization $\upsilon(\hat{\bm{x}}'|\bm{y},\bm{u},\bm{\phi})$,\\
	a memory buffer object containing episodes,\\
	a loss function $\mathcal{L}_\upsilon$,\\
	a learning rate $\alpha_\upsilon$
	\Loop\, for $n_\upsilon$ batches:
	    \State Sample $N_\upsilon$ episodes of length $T$ from memory
	    \State $L \gets 0$ 
	    \Loop\, for every episode $e$ in sample (this is done in parallel):
	        \Loop\, for every step $(\bm{y}, \bm{u}, \bm{y}')$ in $e$:
	            \State Predict next state $\hat{\bm{x}}' = \upsilon(\bm{y},\bm{u},\bm{\phi})$
	            \State Evaluate prediction and update loss $L \gets L + \mathcal{L}_\upsilon(\hat{\bm{x}}', \bm{y}')$
	        \EndLoop
	    \EndLoop
	    \State $\bm{\phi} \gets \bm{\phi} + \alpha_\upsilon \nabla_\phi \frac{L}{TN}$ (using Adam optimizer)
    \EndLoop
    \State Return: $\bm{\phi}$
\end{algorithmic}
\end{algorithm}

\begin{algorithm}[H]
\label{alg:policy_offline}
\caption{Updating of policy network parameters}
\begin{algorithmic}[1]
	\State
	Input: a differentiable transition parametrization $\upsilon(\hat{\bm{x}}'|\bm{y},\bm{u},\bm{\phi})$,\\
	a differentiable policy parametrization $\pi (\bm{u}|\bm{x},\tilde{\bm{x}},\bm{\theta})$,\\
	a memory buffer object containing episodes,\\
	a loss function $\mathcal{L}_\pi$,\\
	a learning rate $\alpha_\pi$,\\
	a number of warm-up steps $w$ and unroll step $r$
	\Loop\, for $n_\pi$ batches:
	    \State Sample $N_\pi$ episodes of length $T$ from memory
	    \State $L \gets 0$ 
	    \State $n_{\text{rollouts}} \gets \lfloor{\frac{T}{w}}\rfloor$ 
	    \Loop\, for every episode $e$ in sample (this is done in parallel):
	        \Loop\, for every rollout in $n_{rollouts}$:
	            \State Reset hidden state of transition and policy networks
	            \State Warm up both models by providing the next $w$ steps $(\bm{y}, \tilde{\bm{x}}, \bm{u}, \bm{y}')$ from $e$
	            \State Predict next state $\hat{\bm{x}}' = \upsilon(\bm{y},\bm{u},\bm{\phi})$
	            \Loop\, for $r$ steps:
    	            \State Predict next hypothetical action $\hat{\bm{u}} = \pi (\hat{\bm{x}}',\tilde{\bm{x}},\bm{\theta})$
    	            \State Predict next hypothetical state $\hat{\bm{x}}' = \upsilon(\bm{y},\bm{\hat{u}},\bm{\phi})$
    	            \State Evaluate hypothetical trajectory and update loss $L \gets L + \mathcal{L}_\pi(\hat{\bm{x}}', \tilde{\bm{x}})$
	            \EndLoop
	        \EndLoop
	    \EndLoop
	    \State $\bm{\theta} \gets \bm{\theta} + \alpha_\pi \nabla_\theta \frac{L}{Nrn_{rollouts}}$ (using Adam optimizer)
    \EndLoop
    \State Return: $\bm{\theta}$
\end{algorithmic}
\end{algorithm}

\vfill
\pagebreak

\section{Training Parameters}
\label{sec:parameters}

The parameters to reproduce the experiments are summarized in \Cref{tab:parameters}. Training was conducted continuously over 1,500 episodes of 4\,s each, making the total exposure to the dynamics to be learned 300,000 steps or 100 minutes. During this process, both models were updated a total of 4,500 times.

\begin{table}[H]
    \caption{Hyperparamters used to obtain data shown in results section.}
    \label{tab:parameters}
    \centering
    \begin{tabular}{p{5cm} S}
        \textbf{Parameter} & \textbf{Value\,\,\,}\\
        \hline
        
        \textit{Task} & \\
        episode steps $T$ & 200\\
        episodes per iteration $E$ & 10\\
        iterations $I$ & 150 \\
        time step [s] $\Delta t$ & 0.02 \\
        memory size $M$ & 1500 \\
        rotation angle [deg] $\gamma$ & 0.0 \\
        acceleration constant $\kappa$ & 5.0 \\
        process noise std. $\zeta$ & 0.001 \\
        observation noise std. $\xi$ & 0.001 \\
        position range $x_{max}$ & 1.0 \\
        velocity range $\dot{x}_{max}$ & 1.0 \\
        control range $u_{max}$ & 1.0 \\
        \hline
        
        \textit{Transition model} & \\
        hidden layer size (MLP) & 256 \\
        learning rate $\alpha_\upsilon$ & 0.0005\\
        batches per iteration $n_\upsilon$ & 30 \\
        batch size $N_\upsilon$ & 1024 \\
        \hline
        
        \textit{Policy model} & \\
        hidden layer size (GRU) & 256 \\
        learning rate $\alpha_\pi$ & 0.0005\\
        batches per iteration $n_\pi$ & 30 \\
        batch size $N_\pi$ & 1024 \\
        warmup steps $w$ & 30\\
        unroll steps $r$ & 20\\
        
    \end{tabular}

\end{table}

\vfill
\pagebreak

\section{LQR Baseline}
\label{sec:lqr}

To compare the learned model with an optimal control theory-based approach, we implemented and hand-tuned a linear quadratic regulator (LQR) \cite{kalman1960contributions}. We used the Python 3 control library for the implementation. The input matrices describe system dynamics $A$, control influence $B$, as well as state cost $Q$ and control cost $R$ and were specified as follows:

\begin{multicols}{2}
\begin{equation*}
\label{eq:a}
\begingroup 
\setlength\arraycolsep{5pt}
    \bm{A} = 
    \begin{bmatrix}
    0 & 0 & 1 & 0 \\
    0 & 0 & 0 & 1 \\
    0 & 0 & 0 & 0 \\
    0 & 0 & 0 & 0
    \end{bmatrix},
\endgroup
\end{equation*}

\begin{equation*}
\label{eq:b}
\begingroup 
\setlength\arraycolsep{5pt}
    \bm{B} = 
    \begin{bmatrix}
    0 & 0 \\
    0 & 0 \\
    \kappa & 0 \\
    0 & \kappa 
    \end{bmatrix},
\endgroup
\end{equation*}
\end{multicols}

\begin{equation}
\label{eq:lqr}
\end{equation}
\vspace{-2cm}

\begin{multicols}{2}
\begin{equation*}
\label{eq:q}
\begingroup 
\setlength\arraycolsep{5pt}
    \bm{Q} = 
    \begin{bmatrix}
    1 & 0 & 0 & 0 \\
    0 & 1 & 0 & 0 \\
    0 & 0 & 0.1 & 0 \\
    0 & 0 & 0 & 0.1
    \end{bmatrix},
\endgroup
\end{equation*}
\break
\begin{equation*}
\label{eq:r}
\begingroup 
\setlength\arraycolsep{5pt}
    \bm{R} = 
    \begin{bmatrix}
    0.1 & 0 \\
    0 & 0.1
    \end{bmatrix}.
\endgroup
\end{equation*}
\end{multicols}

This results in the control gain matrix $K$ as

\begin{equation}
\label{eq:k}
\begingroup 
\setlength\arraycolsep{5pt}
    \bm{K} = 
    \begin{bmatrix}
    3.16227766 & 0.         & 1.50496215 & 0.       \\ 
    0.         & 3.16227766 & 0.         & 1.50496215
    \end{bmatrix}.
\endgroup
\end{equation}

\begin{figure}
    \centering
    \includegraphics[width=.6\textwidth]{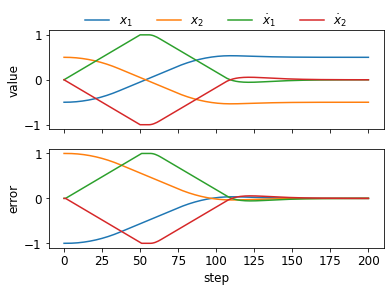}
    \caption{State dynamics under LQR control show that initially, velocity is increased towards the target at the maximum rate, before it plateaus and declines at the same maximum rate. The tuned controller only has minimal overshoot at the target position.}
    \label{fig:lqr}
\end{figure}

Controlling the task described in \Cref{sec:task} to go from the initial state $\bm{x} = [-0.5, 0.5, 0, 0]$ to the target state $\tilde{\bm{x}} = [0.5, -0.5, 0, 0]$ results in the state evolution as shown in \Cref{fig:lqr}.

